\documentclass[lnicst]{svmultln}
\usepackage{makeidx}  

\usepackage{graphics}
\usepackage{graphicx}

\begin{document}
\mainmatter              
\title{Syndromic classification of Twitter messages}
\titlerunning{DIZIE}  
%
\author{Nigel Collier\inst{1} \and Son Doan\inst{1}}
\authorrunning{Nigel Collier et al.}   
%
\tocauthor{Nigel Collier, Son Doan}
\institute{National Institute of Informatics, Tokyo, Japan \\
collier@nii.ac.jp\\
WWW home page: \texttt{http://born.nii.ac.jp/dizie}}

\maketitle              

\begin{abstract}        
Recent studies have shown strong correlation between social networking data and national influenza rates. We expanded upon this success to develop an automated text mining system that classifies Twitter messages in real time into six syndromic categories based on key terms from a public health ontology. 10-fold cross validation tests were used to compare Naive Bayes (NB) and Support Vector Machine (SVM) models on a corpus of 7431 Twitter messages. SVM performed better than NB on 4 out of 6 syndromes. The best performing classifiers showed moderately strong F1 scores: respiratory = 86.2 (NB); gastrointestinal = 85.4 (SVM polynomial kernel degree 2); neurological = 88.6 (SVM polynomial kernel degree 1); rash = 86.0 (SVM polynomial kernel degree 1); constitutional = 89.3 (SVM polynomial kernel degree 1); hemorrhagic = 89.9 (NB). The resulting classifiers were deployed together with an EARS C2 aberration detection algorithm in an experimental online system. 
\end{abstract}
\section{Introduction}
{\it Twitter} is a social networking service that allows users throughout the world to communicate their personal experiences, opinions and questions to each other using micro messages (\lq tweets'). The short message style reduces thought investment \cite{java:2007} and encourages a rapid \lq on the go' style of messaging from mobile devices. Statistics show that Twitter had over 200 million users\footnote{http://www.bbc.co.uk/news/business-12889048} in March 2011, representing a small but significant fraction of the international population across both age and gender\footnote{http://sustainablecitiescollective.com/urbantickurbantick/20462/twitter-usage-view-america} with a bias towards the urban population in their 20s and 30s. Our recent studies into novel health applications \cite{collier:2011c} have shown progress in identifying free-text signals from tweets that allow influenza-like illness (ILI) to be tracked in real time. Similar studies have shown strong correlation with national weekly influenza data from the Centers for Disease Control and Prevention and the United Kingdom's Health Protection Agency. Approaches like these hold out the hope that low cost sensor networks could be deployed as early warning systems to supplement more expensive traditional approaches. Web-based sensor networks might prove to be particularly effective for diseases that have a narrow window for effective intervention such as pandemic influenza.

Despite such progress, studies into deriving linguistic signals that correspond to other major syndromes have been lacking. Unlike ILI, publicly available gold standard data for other classes of conditions such as gastrointestinal or neurological illnesses are not so readily available. Nevertheless, the previous studies suggest that a more comprehensive early warning system based on the same principles and approaches should prove effective. Within the context of the DIZIE project, the contribution of this paper is (a) to present our data classification and collection approaches for building syndromic classifiers; (b) to evaluate machine learning approaches for predicting the classes of unseen Twitter messages; and (c) to show how we deployed the classifiers for detecting disease activity.   A further goal of our work is to test the effectiveness of outbreak detection through geo-temporal aberration detection on aggregations of the classified messages. This work is now ongoing and will be reported elsewhere in a separate study.

\subsection{Automated Web-sensing}

In this section we make a brief survey of recent health surveillance systems that use the Web as a sensor source to detect infectious disease outbreaks. Web reports from news media, blogs, microblogs, discussion forums, digital radio, user search queries etc. are considered useful because of their wide availability, low cost and real time nature. Although we will focus on infectious disease detection it is worth noting that similar approaches can be applied to other public health hazards such as earthquakes and typhoons \cite{earle:2010,sakaki:2010}. 

Current systems fall into two distinct categories: (a) event-based systems that look for direct reports of interest in the news media (see \cite{hartley:2010} for a review), and (b) systems that exploit the human sensor network in sites like Twitter, Jaiku and Prownce by sampling reports of symptoms/GP visits/drug usage etc. from the population at risk \cite{szomszor:2009,lampos:2010,signorini:2011}. Early alerts from such systems are typically used by public health analysts to initiate a risk analysis process involving many other sources such as human networks of expertise.

Work on the analysis of tweets, whilst still a relatively novel information source, is related to a tradition of syndromic surveillance based on analysis of triage chief complaint (TCC) reports, i.e. the initial triage report outlining the reasons for the patient visit to a hospital emergency room. Like tweets they report the patient's symptoms, are usually very brief, often just a few keywords and can be heavily abbreviated. Major technical challenges though do exist: unlike TCC reports tweets contain a very high degree of noise (e.g. spam, opinion, re-tweeting etc.) as well as slang (e.g. {\it itcy} for {\it itchy}) and emoticons which makes them particularly challenging. Social media is inherently an informal medium of communication and lacks a standard vocabulary although Twitter users do make use of an evolving semantic tag set. Both TCC and tweets often consist of short telegraphic statements or ungrammatical sentences which are difficult for uncustomised syntactic parsers to handle.

In the area of TCC reports we note work done by the RODS project \cite{wagner:2004} that developed automatic techniques for classifying reports into a list of syndromic categories based on natural language features. The chief complaint categories used in RODS were respiratory, gastrointestinal, botulinic, constitutional, neurologic, rash, hemorrhagic and none. Further processes took aggregated data and issued alerts using time series aberration detection algorithms.  The DIZIE project which we report here takes a broadly similar approach but applies it to user generated content in the form of Twitter messages.

\section{Method}

DIZIE currently consists of the following components: (1) a list of latitudes and longitudes for target world cities based on Twitter usage; (2) a lexicon of syndromic keywords used as an initial filter, (3) a supervised machine learning model that converts tweets to a word vector representation and then classifies them according to six syndromes, (4) a post-processing list of stop words and phrases that blocks undesired contexts, (5) a MySQL database holding historic counts of positive messages by time and city location, used to calculate alerting baselines, (6) an aberation detection algorithm, and (7) a graphical user interface for displaying alerts and supporting evidence. 

After an initial survey of high frequency Twitter sources by city location we selected 40 world cities as candidates for our surveillance system. Sampling in the runtime system is done using the Twitter API by searching for tweets originating within a 30km radius of a city's latitude and longitude, i.e. a typical commuting/shopping distance from the city centre. The sampling rate is once every hour although this can be shortened when the system is in full operation. In this initial study we focussed only on English language tweets and how to classify them into 6 syndromic categories which we describe below.

Key assumptions in our approach are that: (a) each user is considered to be a sensor in the environment and as such no sensor should have the capacity to over report. We controlled over reporting by simply restricting the maximum number of messages per day to be 5 per user; (b) each user reports on personal observations about themselves or those directly known to them. To control (a) and (b) and prevent over-reporting we had to build in filtering controls to mitigate the effects of information diffusion through re-reporting, particularly for public personalities and mass events. Re-tweets, i.e. repeated messages, and tweets involving external links were automatically removed.

\subsection{Schema development}

A syndrome is a collection of symptoms (both specific and non-specific) agreed by the medical community that are indicative of a class of diseases. We chose six syndrome classes as the targets of our classifier: constitutional, respiratory, gastrointestinal, hemorrhagic, rash (i.e. dermatological) and neurological. These were based on an openly available public health ontology developed as part of the BioCaster project \cite{collier:2008a} by a team of experts in computational linguists, public health, anthropology and genetics. Syndromes within the ontology were based on RODS syndrome definitions and are linked to symptom terms - both technical and laymen's terms - through typed relations. We use these symptoms (syndromic keywords) as the basis for searching Twitter and expanded them using held out Twitter data.

\subsection{Twitter Data}

After defining our syndromes we examined a sample of tweets and wrote guidelines outlining positive and negative case definitions. These guidelines were then used by three student annotators to classify a sample of 2000 tweets per syndrome into positive or negative for each of the syndrome classes. Data for training was collected by automatically searching Twitter using the syndromic keywords over the period 9th to 24th July 2010. No city filtering was applied when we collected the training data. Typical positive example messages are: \lq\lq Woke up with a stomach ache!", \lq\lq Every bone in my body hurts", and \lq\lq Fever, back pain, headache... ugh!". Examples of negative messages are: \lq\lq I'm exercising till I feel dizzy", \lq\lq Cabin fever is severe right now", \lq\lq Utterly exhausted after days of housework". Such negative examples include a variety of polysemous symptom words such as {\it fever} in its senses of raised temperature and excitement and  {\it headache} in its senses of a pain in the head or an inconvenience. The negative examples also include cases where the context indicates that the cause of the syptom is unlikely to be an infection, e.g. headache caused by working or exercising. The training corpus is characterised using the top 7 terms calculated by mutual association score in Table \ref{table2}. This includes several spurious associations such as \lq rt' standing for \lq repeat tweet', \lq botox' which is discussed extensively as a treatment for several symptoms and \lq charice' who is a new pop idol. 

The final corpus was constructed from messages where there was total agreement between all three annotators. This data set was used to develop and evaluate supervised learning classifiers in cross-fold validation experiments. A summary of the data set is shown in Table \ref{table1}. Inter-annotator agreement scores between the three annotators are given as Kappa showing agreement between the two highest agreeing annotators. Kappa indicates strong agreement on most syndromic classes with the noteable exception of gastrointestina and neurological.

\begin{table*}
\centering
\caption{Top 7 terms by syndrome calculated by mutual information score. * indicates a spurious association.}
\begin{tabular}{l l l l l l} 
\hline
Resp   &  Gastro  &  Const & Hemor & Rash & Neuro \\
\hline
throat  & stomach & botox$^{*}$ & pain & road & headache\\
sore &  ache & body & hemorrhage & heat & coma\\
cough & gib & charice$^{*}$ & muscle & arm & worst\\
flu & feel & jaw & tired & tired & gave \\
nose & rt$^{*}$ & hurts & pray & rash & giving \\
rt$^{*}$ & bad & stomach & brain & itcy & vertigo\\
cold & worst & sweating & guiliechelon$^{*}$ & face & pulpo$^{*}$ \\
\hline
\end{tabular}
\label{table2}
\end{table*}

\begin{table*}
\centering
\caption{Structure of the annotated syndrome corpus of Twitter messages.}
\begin{tabular}{l l l l l} 
\hline
Syndrome & Positives (P)  & Negatives (N) & P/N & Kappa\\
\hline
Respiratory & 627 & 738 & 0.85 & 0.67\%\\
Gastrointestinal & 489 & 676 & 0.72 & 0.49\% \\
Neurological & 549 & 434 & 1.26 &  0.42\%\\
Rash & 914 & 592 & 1.54 & 0.86\%\\
Hemorrhagic & 320 & 711 & 0.45 & 0.92\% \\
Constitutional & 1043 & 338 & 3.09 & 0.78\%\\
\hline
\end{tabular}
\label{table1}
\end{table*}

\subsection{Classifier models}

DIZIE employs a two stage filtering process. Since Twitter many topics unrelated to disease outbreaks, DIZIE firstly requests Twitter to send it messages that correspond to a set of core syndromic keywords, i.e. the same sampling strategy used to collect training/testing data. These keywords are defined in the BioCaster public health ontology \cite{collier:2008a}. In the second stage messages which are putatively on topic are filtered more rigorously using a machine learning approach. This stage of filtering aims to identify messages containing ambiguous words whose senses are not relevant to infectious diseases and messages where the cause of the symptoms are not likely to be infectious diseases. About 70\% of messages are removed at this second stage.

To aid in model selection our experiments used two widely known machine learning models to classify Twitter messages into a fixed set of syndromic classes: Naive Bayes (NB) and support vector machines (SVM) \cite{joachims:98} using a variety of kernel functions. Both models were trained with binary feature vectors representing a dictionary index of words in the training corpus. i.e. a feature for the test message was marked 1 if a word was present in the test message which had been seen previously in the training corpus otherwise 0. No normalisation of the surface words was done, e.g. using stemming, because of the high out of vocabulary rate with tools trained on general language texts. 

Despite the implausibility of its strong statistical independence assumption between words, NB tends to perform strongly.  The choice to explore keywords as features rather than more sophisticated parsing and conceptual analysis such as MPLUS \cite{christensen:2002} was taken from a desire to evaluate less expensive approaches before resorting to time consuming knowledge engineering. 

The NB classifier exploits an estimation of the Bayes Rule:

\begin{equation}
P(c_{k}|d) = \frac{P(c_{k}) \times \prod_{i=1}^{m}P(f_{i}|c_{k})^{f_{i}(d)}}{P(d)}
\end{equation}

where the objective is to assign a given feature vector for a document $d$ consisting of $m$ features to the highest probability class $c_{k}$. $f_{i}(d)$ denotes the frequency count of feature $i$ in document $d$. Typically the denominator $P(d)$ is not computed explicitly as it remains constant for all $c_{k}$. In order to compute the highest value numerator NB makes an assumption that features are conditionally independent given the set of classes. Right hand side values of the equation are estimates based on counts observed in the training corpus of classified Twitter messages. We used the freely available Rainbow toolkit\footnote{http://www.cs.cmu.edu/~mccallum/bow/rainbow/} from CMU as the software package.

SVMs have been widely used in text classification achieving state of the art predictive accuracy. The major distinction between the two approaches are that whereas NB is a generative classifier which forms a statistical model of each class, SVM is a large-margin binary classifier. SVM operates as a two stage process. Firstly the feature vectors are projected into a high dimensional space using a kernel function. The second stage finds a maximum margin hyperplane within this space that separates the positive from the negative instances of the syndromic class. In practice it is not necessary to perfectly classify all instances with the level of tolerance for misclassification being controlled by the C parameter in the model. A series of binary classifiers were constructed (one for each syndrome) using the SVM$^{Light}$ software package \footnote{http://svmlight.joachims.org/}. We explored polynomial degree 1, 2, 3 and radial basis function kernels.

\subsection{Temporal model}

In order to detect unexpected rises in the stream of messages for each syndrome we implemented a widely used change point detection algorithm called the Early Aberration and Reporting System (EARS) C2 \cite{hutwagner:2003}.  C2 reports an alert when its test value $S_{t}$ exceeds a number $k$ of standard deviations above a historic mean:

\begin{equation}
S_{t} = max(0,(C_{t} - (\mu_{t} + k\sigma_{t}))/\sigma_{t})
\end{equation}

where $C_{t}$ is the count of classified tweets for the day, $\mu_{t}$ and $\sigma_{t}$  are the mean and standard deviation of the counts during the history period, set as the previous two weeks. $k$ controls the number of standard deviations above the mean where an alert is triggered, set to 1 in our system. The output of C2 is a numeric score indicating the degree of abnormality but this by itself is not so meaningful to ordinary users. We constructed 5 banding groups for the score and showed this in the graphical user interface.

\section{Results}

\subsection{Classifying Twitter Messages}

Results for 10-fold cross validation experiments on the classification models are shown in Table  \ref{table3}. Overall the SVM with polynomial degree 1 kernel outperformed all other kernels with other kernels generally offering better precision at a higher cost to recall. Precision (Positive predictive) values ranged from 82.0 to 93.8 for SVM (polynomial degree 1) and from 83.3 to 99.0 for NB. Recall (sensitivity) values ranged from 58.3 to 96.2 for SVM (polynomial degree 1) and from 74.7 to 90.3 for NB. SVM tended to offer a reduced level of precision but better recall.  In the case of one syndrome (Hemorrhagic) we noticed an unusually low level of recall for SVM but not for NB.  

SVM's performance seemed moderately correlated to the positive/negative ratio in the training corpus and also showed weakness for the two classes (Hemorrhagic and Gastrointestinal) with the smallest positive counts. Naive Bayes performed robustly across classes with no obvious correlation either to positive/negative ratio or the volume of training data. Low performance was seen in both models for the gastrointestinal syndrome. This was probably due to the low number of training examples resulting from the low inter-annotator agreement on this class and the requirement for complete agreement between all three annotators.

\begin{table*}
\centering
\caption{Evaluation of automated syndrome classification using naive Bayes and Support Vector Machine models on 10-fold cross validation. P - Precision, R - Recall, F1 - F1 score. $^{1}$ SVM using a linear kernel, $^{2}$ SVM using a polynomial kernal degree 2, $^{3}$ SVM using a polynomial kernal degree 3, $^{R}$ SVM using a radial basis function kernel.}
\begin{tabular}{l l l l l l l l l l l l l l l l} 
\hline
& \multicolumn{3}{c}{Naive Bayes} & \multicolumn{3}{c}{SVM$^{1}$} & \multicolumn{3}{c}{SVM$^{2}$} & \multicolumn{3}{c}{SVM$^{3}$} & \multicolumn{3}{c}{SVM$^{R}$} \\
\hline
Synd. & P & R & F1 & P & R & F1 & P & R & F1 & P & R & F1 & P & R & F1\\
\hline
Resp. & 90.3 & 82.4 & 86.2 & 85.4 & 82.5 & 83.8 & 83.0 & 71.0 & 76.5 & 86.4 & 61.3 & 71.7 & 66.7 & 3.2 & 6.2\\
Gast. & 83.3 & 75.5 & 79.2 & 85.9 & 78.4 & 81.8 & 92.7 & 79.2 & 85.4 & 91.4 & 66.7 & 77.1 & 73.1 & 39.6 & 51.3 \\
Neur. & 98.2 & 74.7 & 84.8 & 83.2 & 95.0 & 88.6 & 77.9 & 98.2 & 86.9 & 62.4 & 98.2 & 76.3 & 90.0 & 63.0 & 74.1 \\
Rash & 94.5 & 76.1 & 84.3 & 82.0 & 90.6 & 86.0 & 76.9 & 91.2 & 83.4 & 67.7 & 94.5 & 78.9 & 60.7 & 100.0 & 75.5\\
Hem. & 89.4 & 90.3 & 89.9 & 93.8 & 58.3 & 71.7 & 100.0 & 50.0 & 66.7 & 100.0 & 50 & 66.7 & 87.5 & 43.8 & 58.3\\
Con. & 99.0 & 79.8 & 88.4 & 83.6 & 96.2 & 89.3 & 83.6 & 93.3 & 88.2 & 78.6 & 99.0 & 87.7 & 76.5 & 100 & 86.7\\
\hline
\end{tabular}
\label{table3}
\end{table*}

\subsection{Technology dissemination}

\begin{figure*}[t]
\label{fig1}
\begin{center}
\includegraphics[bb=270 0 380 650,scale=0.3]{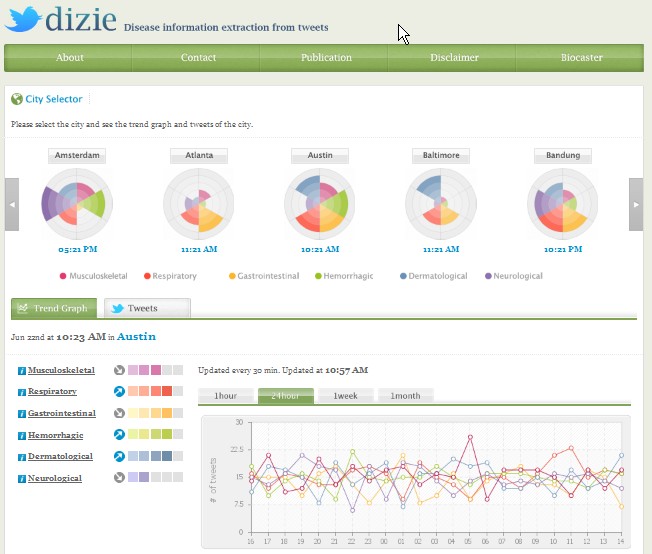}
\caption{Radial graphs showing syndromic alert levels for major world cities. Colour coding on the radial segments indicates the alerting degree automatically assigned to a syndrome in a city based on the previous hour's Twitter counts and the previous 2 weeks as a baseline. The page is updated every hour. Clicking on the graph for a city displays the frequency graph and also the matching tweets for the current hour.}
\end{center}
\end{figure*}

\begin{figure*}[t]
\label{fig2}
\begin{center}
\includegraphics[bb=0 0 350 250,scale=0.6]{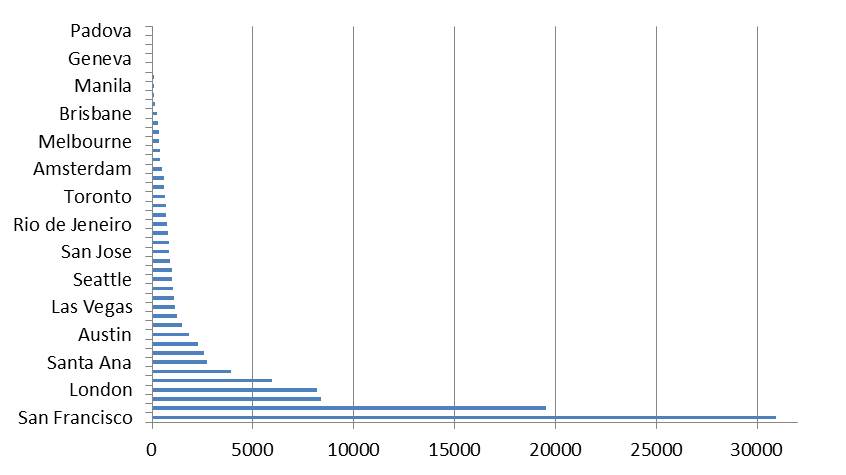}
\caption{Number of Tweets by a sample of major world cities classified by DIZIE during the period 2nd March 2011 to 31st August 2011.}
\end{center}
\end{figure*}

An experimental service for syndromic surveillance called DIZIE has been implemented based on the best of our classifier models and we are now observing its performance. The service is freely available from an online portal at http://born.nii.ac.jp/dizie. As shown in Figure \ref{fig1} the graphical user interface (GUI) for DIZIE shows a series of radial charts for each major world city with each band of the chart indicating the current level of alert for one of the six syndromes. Alerting level scores are calculated using the Temporal Model presented above. Each band is colour coded for easy recognition. Alerting levels are calculated on the classified twitter messages using the EARS C2 algorithm described above. Data selection is by city and time with drill down to a selection of user messages that contributed to the current level. Trend bars show the level of alert and whether the trend is upwards, downwards or sideways. Charting is also provided over an hourly, daily, weekly and monthly period. The number of positively classified messages by city is indicated in Figure \ref{fig2} for a selection of cities. 

Navigation links are provided to and from BioCaster, a news event alerting system, and we expect in the future to integrate the two systems more closely to promote greater situation awareness across media sources.  Access to the GUI is via regular Web browser or mobile device with the page adjusting automatically to fit smaller screens.

\section{Conclusion}

Twitter offers unique challenges and opportunities for syndromic surveillance. Approaches based on machine learning need to be able (a) to handle biased data, and (b) to adjust to the rapidly changing vocabulary to prevent a flood of false positives when new topics trend. Future work will compare keyword classifiers against more conceptual approaches such as \cite{christensen:2002} and also compare the performance characteristics of change point detection algorithms. 

Based on the experiments reported here we have built an experimental application called DIZIE that samples Twitter messages originating in major world cities and automatically classifies them according to syndromes. Access to the system is openly available. Based on the outcome of our follow up study we intend to integrate DIZIE's output with our event-based surveillance system BioCaster which is currently used by the international public health community.

\section*{Acknowledgements}

This work was in part supported by grant in aid support from the National Institute of Informatics' Grand Challenge Project (PI:NC). We are grateful to Reiko Matsuda Goodwin for commenting on the user interface in the early stages of this study and helping in data collection for the final system.

\bibliographystyle{unsrt}

\bibliography{../../../Thesis/report,../../../Thesis/thesis,../../../Thesis/mypublications}

\begin{thebibliography}{10}

\bibitem{java:2007}
A.~{Java}, X.~{Song}, T.~{Finin}, and B.~{Tseng}.
\newblock Why we twitter: Understanding microblogging usage and communities.
\newblock In {\em Proc. 9th {WebKDD} and 1st {SNA-KDD} Workshop on Web Mining
  and Social Network Analysis, ACM}, 12th August 2007.

\bibitem{collier:2011c}
N.~{Collier}, S.~T. {Nguyen}, and M.T.N. {Nguyen}.
\newblock {OMG U got flu?} analysis of shared health messages for
  bio-surveillance.
\newblock {\em Biomedical Semantics}, 2(Suppl 5):S10, September 2011.

\bibitem{earle:2010}
P.~{Earle}.
\newblock Earthquake twitter.
\newblock {\em Nature Geoscience}, 3(4):221--222, 2010.
\newblock doi:10.1038/ngeo832.

\bibitem{sakaki:2010}
T.~{Sakaki}, M.~{Okazaki}, and Y.~{Matsuo}.
\newblock Earthquake shakes twitter users: real-time event detection by social
  sensors.
\newblock In {\em Proc. of the 19th International World Wide Web Conference,
  Raleigh, NC, USA}, pages 851--860, 2010.

\bibitem{hartley:2010}
D.~{Hartley}, N.~{Nelson}, R.~{Walters}, R.~{Arthur}, R.~{Yangarber},
  L.~{Madoff}, J.~{Linge}, A.~{Mawudeku}, N.~{Collier}, J.~{Brownstein},
  G.~{Thinus}, and N.~{Lightfoot}.
\newblock The landscape of international biosurveillance.
\newblock {\em Emerging Health Threats J.}, 3(e3), January 2010.
\newblock doi:10.1093/bioinformatics/btn534.

\bibitem{szomszor:2009}
Martin Szomszor, Patty Kostkova, and Ed~De Quincey.
\newblock {\em swineflu : Twitter predicts swine flu outbreak in 2009}.
\newblock Number December. 2009.

\bibitem{lampos:2010}
V.~{Lampos}, T.~{De Bie}, and N.~{Cristianini}.
\newblock Flu detector - tracking epidemics on twitter.
\newblock In {\em Machine Learning and Knowledge Discovery in Databases},
  volume 6223/2010, pages 599--602. Lecture Notes in Computer Science, 2010.

\bibitem{signorini:2011}
A.~{Signorini}, A.~M. {Segre}, and P.~M. {Polgreen}.
\newblock The use of twitter to track levels of disease activity and public
  concern in the {U.S.} during the influenza a h1n1 pandemic.
\newblock {\em {PLoS One}}, 6(5):e19467, 2011.

\bibitem{wagner:2004}
M.~M. {Wagner}, J.~{Espino}, F.C. {Tsui}, P.~{Gesteland}, W.~{Chapman},
  W.~{Ivanov}, A.~{Moore}, W.~{Wong}, J.~{Dowling}, and J.~{Hutman}.
\newblock Syndrome and outbreak detection using chief-complaint data -
  experience of the real-time outbreak and disease surveillance project.
\newblock {\em Morbidity and Mortality Weekly Report (MMWR)}, 53
  (Suppl):28--31, 2004.

\bibitem{collier:2008a}
N.~{Collier}, S.~{Doan}, A.~{Kawazoe}, R.~{Matsuda Goodwin}, M.~{Conway},
  Y.~{Tateno}, Q.~{Ngo}, D.~{Dien}, A.~{Kawtrakul}, K.~{Takeuchi},
  M.~{Shigematsu}, and K.~{Taniguchi}.
\newblock {BioCaster}:detecting public health rumors with a web-based text
  mining system.
\newblock {\em Bioinformatics}, 24(24):2940--1, December 2008.
\newblock doi:10.1093/bioinformatics/btn534.

\bibitem{joachims:98}
T.~{Joachims}.
\newblock Text categorization with support vector machines: Learning with many
  relevant features.
\newblock In {\em Proceedings of the European Conference on Machine Learning},
  1998.

\bibitem{christensen:2002}
L.~M. {Christensen}, P.~J. {Haug}, and M.~{Fiszmann}.
\newblock Mplus: A probabilistic medical language understanding model.
\newblock In {\em Proceedings of the Workshop on Natural Language Processing in
  the Biomedical Domain, Philadelphia, USA}, July 2002.

\bibitem{hutwagner:2003}
L.~{Hutwagner}, W.~{Thompson}, M.~G. {Seeman}, and T.~{Treadwell}.
\newblock The bioterrorism preparedness and response early aberration reporting
  system ({EARS}).
\newblock {\em J. Urban Health}, 80(2):i89--i96, 2003.

\end{thebibliography}

\end{document}